\pdfoutput=1

\documentclass[final,authoryear,5p,times,twocolumn]{elsarticle}

\usepackage{graphicx}

\usepackage{algorithm}
\usepackage{algorithmic}

\usepackage{amssymb}
\usepackage{amsmath}
\usepackage{amsthm}
\usepackage{url}

\newdefinition{definition}{Definition}

\journal{Knowledge-Based Systems}

\begin{document}

\begin{frontmatter}



\title{Dyna-$\mathcal{H}$: a heuristic planning reinforcement learning algorithm applied to role-playing game strategy decision systems}


\author{Matilde Santos} \ead{msantos@dacya.ucm.es}
\author{Jos\'e Antonio Mart\'in H.\corref{cor1}} \ead{jamartinh@fdi.ucm.es}
\author{Victoria L\'opez} \ead{vlopez@dacya.ucm.es}
\author{Guillermo Botella} \ead{gbotella@fdi.ucm.es}
\cortext[cor1]{address: Facultad de Inform\'atica, Universidad Complutense de Madrid,
C. Prof. Jos\'e Garc\'ia Santesmases, s/n., 28040, Madrid, (Spain). Tel.: +34 91.394.7620, Fax: +34  91.394.7510}

\address{Computer Architectures and Automation, Complutense University of Madrid, Spain.}

\begin{abstract}
In a Role-Playing Game, finding optimal trajectories is one of the most important tasks. In fact, the strategy decision system becomes a key component of a game engine. Determining the way in which decisions are taken (online, batch or simulated) and the consumed resources in decision making (e.g. execution time, memory) will influence, in mayor degree, the game performance. When classical search algorithms such as $A^*$ can be used, they are the very first option. Nevertheless, such methods rely on precise and complete models of the search space, and there are many interesting scenarios where their application is not possible. Then, model free methods for sequential decision making under uncertainty are the best choice. In this paper, we propose a heuristic planning strategy to incorporate the ability of heuristic-search in path-finding into a Dyna agent. The proposed Dyna-$\mathcal{H}$ algorithm, as $A^*$ does, selects branches more likely to produce outcomes than other branches. Besides, it has the advantages of being a model-free online reinforcement learning algorithm. The proposal was evaluated against the one-step $Q$-Learning and Dyna-$Q$ algorithms obtaining excellent experimental results: Dyna-$\mathcal{H}$ significatively overcomes both methods in all experiments. We suggest also, a functional analogy between the proposed sampling from worst trajectories heuristic and the role of dreams (e.g. nightmares) in human behavior.
\end{abstract}


\end{frontmatter}

\section{Introduction}
Decision support systems (DSS) are computer-based information systems that support business or any other organizational decision-making activities. DSSs help to make decisions, which may be rapidly changing and not easily specified in advance. DSSs include knowledge-based systems. The importance of making a good decision in any business is evident. In a dynamic environment, decision processes not only need to be well designed but they must adapt rapidly to changes in the environment. Existing work on decision making has centered on the concepts of rational and boundedly rational decision processes. Recent works include a third model of decision, based on the use of heuristics.


In the last years, there has been an increasing interest in the issues of cost-sensitive learning and decision making, in a variety of applications, in order to maximize the total benefits over time. A number of approaches have been developed that are effective at optimizing cost-sensitive decisions~\citep{lopez2010,Iglesias2008}, some of them based on a synergy between different intelligent techniques and other fields that together comprise what is called knowledge engineering~\citep{Lu2007}.

In any decision making strategy, an agent seeks to achieve a goal, despite uncertainty about its environment. The agent's actions influence the future state of the environment, thereby affecting the options and possible alternatives at later times. Correct choice requires taking into account indirect, delayed consequences of actions, and thus may include foresight or planning~\citep{RL98}.

Among all the decisions involved in computer-games, the most common is probably path-finding, i.e., looking for a good route or path for moving an entity from here to there. The entity can be a single person, a vehicle, or a combat unit; the genre can be an action game, a simulator, a role-playing game, or a strategy game. The main focus of this research is to compute collision-free shortest-paths as quickly as possible. Although path-finding is not trivial, there are some well-established, solid algorithms that have been applied, some of them more efficient than others~\citep{Bayili2011,Alvarez2010}.

In this paper we use, as the case study, the Role-Playing Games (RPG) scenario, where the player selects a target point $(t)$ from its current position and the entity $(e)$ is automatically taken to $t$ without interacting with the system, avoiding obstacles and optimizing the trajectory. This automatic process can be carried out by different approaches~\citep{Karamouzas2008}. Most of the searching strategies proposed in the literature are included in the wide area of machine learning~\citep{Alpaydin2004,mitchell97}. When classical search algorithms such as $A^*$ can be used, they are the very first choice for computing optimal solutions. Nevertheless, these methods can be computationally demanding, especially for very large environments. For instance, $A^*$ based algorithms usually demands quite high execution time since the decisions rely on a exhaustive planning strategy. Even more, such methods depend heavily on precise and complete models of the environment, e.g. the game arena. So, there are many interesting scenarios where they cannot be applied. Therefore, model free methods for sequential decision making problems under uncertainty are well suited to these cases since the incremental nature of its learning mechanisms and the direct action selection mechanism of its decision making procedures make it possible to use them in real-time applications.

Many other applications of these learning strategies can be found in the literature. Without being exhaustive, some recent paradigmatic examples can be cited. The airline ticket purchasing problem~\citep{Gilmore2008}, where author uses different techniques to acquire a flight ticket at the lowest cost. MALADY: A Machine Learning-Based Autonomous Decision-Making System for Sensor Networks~\citep{Krishnamurthy2009}, where sensor networks are able to learn and make decisions in real time. \citet{Muse06} present a system for visual robotic docking using an omnidirectional camera coupled with the actor critic reinforcement learning algorithm. In this case, a network trained via reinforcement allows the robot to turn to and approach a table to pick an object. \citet{Janssens2007} present an application of reinforcement learning ($Q$-learning) that simulates time and location information for a given sequence of travel activities. Even in a different field such as education we can find some interesting applications~\citep{Iglesias2009}. In this paper the process of learning pedagogical policies according to the students needs fits an RL problem. \citet{Kaelbling96} and~\citet{Busoniu2008} have written surveys on reinforcement learning and its applications. A heuristic method can use searching trees. However, instead of generating all possible solution branches, a heuristic selects branches more likely to produce successful outcomes than other branches. It is selective at each decision point. This paper is an extension of a previous one on path-finding for RPGs~\citep{Alvarez2010}.

In this article, we introduce a novel algorithm that includes a heuristic planning module (sampling from the worst trajectories) and a function $\mathcal{H}$ (the a priori knowledge injected to the system) that can contain any kind of information that express how bad is taking an action at a particular situation, for example, the Euclidean distance between a goal state and the current state. The proposed Dyna-$\mathcal{H}$ algorithm is based on the well-known Dyna architecture~\citep{Sutton91,RL98}.

Grid world like environments treated as Markov sequential decision problems are used nowadays in many research works to evaluate and show the behavior of standard algorithms against new proposed ones. The results obtained in this test cases are easily generalizable to other problems, such as robot navigation, and, in general, any sequential decision problem. In this particular case, to an informed (i.e. knowledge-based) sequential decision problem. The proposed method, the one-step $Q$-learning and Dyna-$Q$ algorithms have been applied to the same problem and compared in terms of learning rate.

The structure of the paper is as follows. In section 2, the strategies that are going to be applied and compared are briefly described and the novel proposal is introduced. The experimental scenario is described in Section 3. Results obtained by the different algorithms are discussed in Section 4. The last section (5) is devoted to the conclusions and further work.

\section{Search, Reinforcement learning and Planning}
The algorithms that are going to be compared are briefly described in this section. A new algorithm based on the Dyna architecture~\citep{Sutton91,RL98}, that combines heuristic on-line search and $Q$-Learning is presented. We focus on solving path planning problems for homogeneous agents in homogeneous environments.

\subsection{Heuristic search, the $A^*$ algorithm}

The predominant state-space planning methods in artificial intelligence are collectively known as heuristic search. Unlike other planning methods, heuristic search is not concerned with changing the approximate, value function, but only with improving the actions selection given the current value function.

In heuristic search, for each state encountered, a large tree of possible alternatives is considered. The approximate value function is applied to the leaf nodes, and then backed up at the previous state towards the root. The backing up in the search tree is just the same as in the max-backups. This backing up stops at the state-action nodes of the current state. Once the backed-up values of these nodes are computed, the best of them is chosen as the current action, and the rest of the values are discarded. In conventional heuristic search no effort is made to save the backed-up values and the value function, once designed, never changes as a result of the search. However, it would be reasonable to allow the value function to be improved over time, using either the backed-up values computed during the heuristic search or by any other method.

Heuristic methods such as $A^*$ based algorithms have been widely applied. Actually, in the game development community, the most popular path-planning is to divide the environment into a grid that can be explored using these $A^*$ based algorithms. This approach works very well in computer games as it always retrieves the shortest path, if exists. This heuristic search ranks each node by an estimate of the best route through that node. It combines the tracking of the previous path length of Dijkstra's algorithm~\citep{Dijkstra59}, with the heuristic estimate of the remaining path from best-first search. Since some nodes may be processed more than once, in order to find better paths later, it is necessary to keep track of them in a list. Adding this heuristic score to the nodes stored in Dijkstra's priority queue, the number of nodes visited during the search can be effectively pruned down.

$A^*$ has a couple of interesting properties. It is guaranteed to find the shortest path, as long as the heuristic estimate is admissible. That is, it is never greater than the remaining distance to the goal. It makes the \emph{most efficient} use of the heuristic function:\emph{ no search that uses the same heuristic function and finds optimal paths will expand fewer nodes than} $A^*$, not counting tie-breaking among nodes of equal cost. $A^*$ turns out to be very flexible in practice.

\subsection{Reinforcement Learning (RL)}

Reinforcement Learning~\citep{Kaelbling96,RL98} goes back to the very first stages of Artificial Intelligence and Machine Learning, and it has several applications in the Intelligent Knowledge Engineering Systems domain. They have been also successfully applied to game playing~\citep{Littman94}.

Under a constrained environment, the learning agent can perceive a set $S$ of distinct states, which are normally characterized by a number of dimensions, and it has a set $A$ of possible actions at each state. Reinforcement learning tasks are generally discrete. At each time step $t$, the agent observes the current state $s_t$ and chooses a possible action $a_t$, which leads to the succeeding state $s_{t+1} = d(s_t,a_t)$. Then, the environment generates a reward $r(s_t,a_t)$. These rewards can be positive, zero or negative and can have a delay. In other words, some actions and their state transitions may bring low rewards in short term, while they will lead to state-action pairs with a much higher reward later. On the contrary, an action in a given state may receive an immediate high reward, whereas it makes the agent enter into a path where the following actions have very low or even negative rewards. Therefore, the task of the agent is to learn a policy $\pi: S \rightarrow A$, to achieve the maximum accumulative reward over time.

Reinforcement learning agents are connected to the environment by perceptions and actions. On each step of the interaction with the environment, the agent receives as input the current state and the value of that state. This value is the reward. The agent records the reward signal and updates the policy based on the information received about the reward so far.

$Q$-Learning~\citep{Watkins92} is a popular method of model-free reinforcement learning. It can also be viewed as a method of asynchronous dynamic programming (DP)~\citep{Bellman57,Bellman62}. Reinforcement Learning provides agents with the capability of learning from interactions with the environment, to act optimally in Markovian domains by experiencing the consequences of actions, without requiring them to receive or build maps (models) of the domains~\citep{Grzes2010}.

Learning proceeds similarly to Sutton's method of temporal differences (TD)~\citep{RL98}: an agent tries an action at a particular state, and evaluates its consequences in terms of the immediate reward or penalty it receives and its estimate of the value of the state to which is taken. By trying all actions in all states repeatedly, it learns which ones are the best overall, judged by long-term discounted cumulative reward~\citep{Tesauro92}.

A probabilistic approach is commonly used in $Q$-learning. A straightforward strategy is the $\epsilon$-greedy method, where the probability of making a random choice is handled by the parameter $\epsilon$. In every step, with probability $1-\epsilon$, the agent fully exploits the information stored in the $Q$-values, and with probability $\epsilon$ the agent chooses a random action in order to explore the state space. In the exploration mode, the $\epsilon$-greedy method assumes equal selection probabilities for any possible action, whereas the chance of selecting a better action may be increased by taking the current value distribution between alternatives such as in the soft-max methods~\citep{RL98}.

\subsection{The Dyna architecture}

Planning is usually referred to any computational process that takes a model as input and produces or improves a policy to interact with the modeled environment. Although there are different approaches, state-space planning is mainly a search through the state space for an optimal path. Actions cause transitions from one state to another, and value functions are computed over states.

\begin{figure}[tb]
  \includegraphics[scale=0.8]{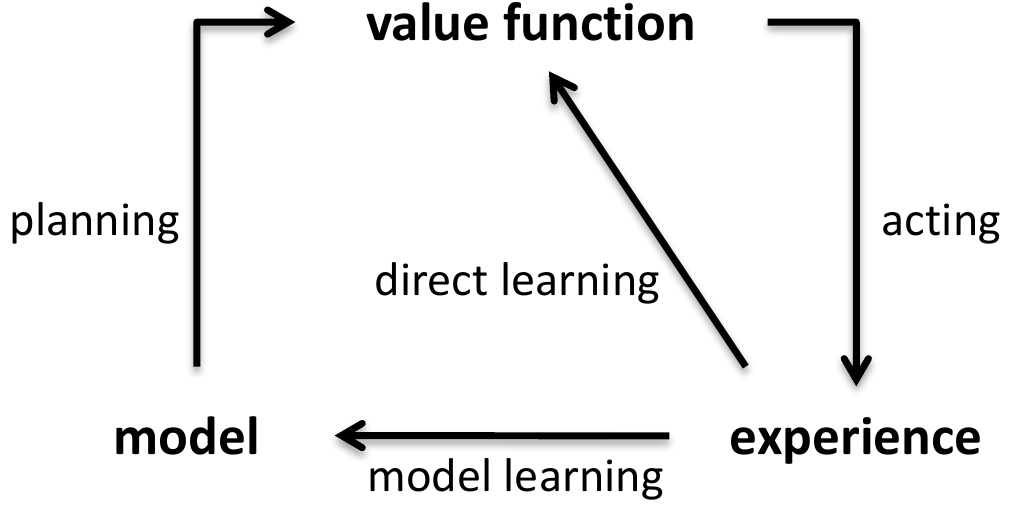}
  \caption{Information flow in the Dyna architecture}\label{dyna-model}
\end{figure}

\begin{algorithm}[tb]
   \caption{Dyna-$Q$ algorithm, as proposed by \citet{Sutton91}.}
   \label{alg:dyna-q}
\begin{algorithmic}[1]
   \STATE Initialize $Q(s,a)$, ${Model}(s,a)$ $\forall$ $s\in \mathcal{S}, \;a\in \mathcal{A}$
   \REPEAT[for each episode]
   \STATE $s \leftarrow$ current(non terminal) state
   \STATE $a \leftarrow \epsilon$-${greedy}(s,Q)$
   \STATE execute $a$; observe $s'$ and $r$
   \STATE $Q(s,a) \leftarrow Q(s,a) + \alpha [r + \gamma \max_{a'} Q(s',a') - Q(s,a)]$
   \STATE  ${Model}(s,a) \leftarrow s',r$
   \FOR{$i = 1$ \TO $N$}
   \STATE $s \leftarrow$ random previously observed state
   \STATE $a \leftarrow$ random action previously taken in $s$
   \STATE $s',r \leftarrow {Model}(s,a)$
   \STATE $Q(s,a) \leftarrow Q(s,a) + \alpha [r + \gamma \max_{a'} Q(s',a') - Q(s,a)]$
   \ENDFOR
   \UNTIL{$s'$ is terminal}
\end{algorithmic}
\end{algorithm}

In on-line planning, new information is gained from the interaction with the environment and may change the model. If decision-making and model-learning are both computation-intensive processes, it may be necessary to divide the available computational resources between them. Dyna~\citep{Sutton91}, is a reinforcement learning architecture that easily integrates incremental reinforcement learning and on-line planning.

The possible relationship between experience, model and values for Dyna-$Q$ are described in figure~\ref{dyna-model}. Each arrow shows a relationship of influence. Note how experience can improve the model and therefore the value function, either directly or indirectly. It is the latter, which is sometimes called indirect reinforcement learning, which is involved in planning. In algorithm~\ref{alg:dyna-q}, where Dyna-$Q$ is described, ${Model}(s,a)$ denotes the contents of the model (predicted next state and reward, respectively) for state-action pair $(s, a)$. Direct reinforcement learning, model-learning, and planning are implemented by steps 6, 7 and 8, respectively. If steps 7 and 8 were omitted, the remaining algorithm would be one-step tabular $Q$-learning.

Dyna-$Q$ includes all of these processes: planning, acting, model-learning, and direct RL, continually. The planning method is the random-sample one-step $Q$-planning. The direct RL method is the one-step $Q$-learning. The model-learning method is table-based and assumes the world is deterministic. After each transition, the model records the prediction that will deterministically follow. Thus, if the model is queried with a state-action pair that has been experienced before, it simply returns the last-observed next state and next reward as its prediction. During planning, the algorithm randomly samples only of state-action pairs that have been previously experienced. Conceptually, planning, acting, model-learning, and direct RL occur simultaneously and in parallel in Dyna agents~\citep{RL98}.

\section{A heuristic planning reinforcement learning algorithm based on the Dyna architecture}

Here we propose a heuristic planning strategy to incorporate into a Dyna agent the advantages of a particular heuristic, in order to find the shortest paths in grid like environments, e.g. RPGs. A heuristic search method, as a search method after all, can be defined in terms of traversing a search tree. However, instead of generating all possible solution branches, a heuristic method selects branches more likely to produce successful outcomes than others. It is selective at each decision point. The proposed method incorporates the ability of heuristic search, e.g. $A^*$, to focus on specific search subtrees in order to make the searching more efficient. At the same time, the method learns online as any other common reinforcement learning algorithm and does not requieres a complete model of the environment before staring to search.

\subsection{Sampling from the worst trajectories (the nightmares metaphor)}

Contrary to intuition, the proposed sampling strategy consist in using a learned model of the environment and traveling across it using the worst trajectories with respect to some heuristic index (e.g. a priori knowledge of the domain), receiving thus the worst rewards. However, this lead the algorithm to find the solution faster that using any other a priori better approach.

Sampling from ``bad'' trajectories using simulated experience has a very interesting analogous in human behavior: nightmares. This analogy suggests that such strategy can be considered as an interesting candidate hypothesis about the role of nightmares in human behavior, assigning thus a specific function to this behavior: a tool used by our brain to reorganize some goal oriented behaviors using the resting time to learn based on imagination (simulated experience). Furthermore, Figure~\ref{trajectories} (in section~\ref{sec:experimental}) show different trajectories using this sampling strategy. As can be seen, these trajectories present some discontinuities (abrupt jumps) and also pass through the walls, i.e. violates the physical laws; things that are very common in dreams.

The analogous heuristic, in this case, to the $\mathcal{H}$ function, could be associated with the so called value-systems, which shape human behavior~\citep{e-ndtngs-87,Sporns2000}. Indeed, there is a growing body of research about value-systems in robotics and autonomous agents in order to design robots with adaptive, lifelong learning behavior, because this values-systems are a way for robots to behave autonomously through spontaneous, self-generated activity~\citep{Merrick2010}. In connection with autonomous agents many kinds of different value-systems, based on some aspects of human behavior related to motivation, e.g. curiosity driven, intrinsic motivated, novelty detection, have been proposed. However, it seems that there is (up to our knowledge) no publication along this line of research relating the study of dreams and value-systems with the Reinforcement Learning and Planning field.

\subsection{The Dyna-$\mathcal{H}$ algorithm}

In RPGs and grid world like environments in general, it is common to use the Euclidian or city-clock distance functions as an effective heuristic. In this case study, the euclidian distance is used for the heuristic $(\mathcal{H})$ planning module. However, in general, $\mathcal{H}(s,a)$ represents a general function that gives a guess about \emph{how bad} is to take action $a$ in state $s$, e.g. the euclidian distance between the state $s'$ and the ${goal}$ position~(\ref{eq:H}).

\begin{equation}
\label{eq:H}
\mathcal{H}(s,a) = ||s'-{goal}||^2, \
\end{equation}
where the $s'$ state is the result of the model query: $s' = {Model}(s,a) $.

Hence, given the heuristic $\mathcal{H}$, the heuristic action ${h_a}$ is defined as:
\begin{equation}
\label{bestaction} h_a(s,\mathcal{H})=\underset{a}{argmax} \; \mathcal{H}(s,a),
\end{equation}
where $h_a(s,\mathcal{H})$ is the worst action following $(\mathcal{H})$, e.g. the action that yields the higher distance from the goal. Algorithm~\ref{alg:dyna-h} describes the steps of this strategy.

\begin{algorithm}
   \caption{The proposed Dyna-$\mathcal{H}$ heuristic planning algorithm}
    \label{alg:dyna-h}
\begin{algorithmic}[1]
   \STATE Initialize $Q(s,a)$, ${Model}(s,a)$ $\forall$ $s\in \mathcal{S}, \;a\in \mathcal{A}$
   \REPEAT[for each episode]
   \STATE $s \leftarrow$ current(non terminal) state
   \STATE $a \leftarrow \epsilon$-${greedy}(s,Q)$
   \STATE execute $a$; observe $s'$ and $r$
   \STATE $Q(s,a) \leftarrow Q(s,a) + \alpha [r + \gamma \max_{a'} Q(s',a') - Q(s,a)]$
   \STATE  ${Model}(s,a) \leftarrow s',r$
   \FOR{$i = 1$ \TO $N$}
   \STATE $a \leftarrow \; h_a(s,\mathcal{H})$
   \IF{$s,a \not \in {Model}$}
   \STATE $s \leftarrow$ random previously observed state
   \STATE $a \leftarrow$ random action previously taken in $s$
   \ENDIF
   \STATE $s',r \leftarrow {Model}(s,a)$
   \STATE $Q(s,a) \leftarrow Q(s,a) + \alpha [r + \gamma \max_{a'} Q(s',a') - Q(s,a)]$
   \STATE $s \leftarrow s'$
   \ENDFOR
   \UNTIL{$s'$ is terminal}
\end{algorithmic}
\end{algorithm}

\section{Experimental scenario}
\label{sec:experimental}
\begin{figure}[b]
  \center
  \includegraphics[scale=0.7]{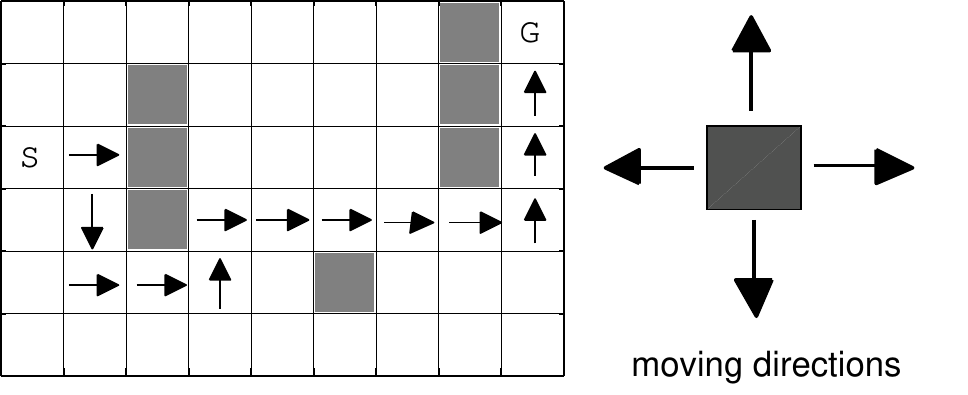}
  \caption{The experimental scenario, starting point $(S)$, goal $(G)$, obstacles (gray), and a sample trajectory.}\label{fig:scenario}
\end{figure}

The Dyna-$\mathcal{H}$ heuristic planning algorithm have been evaluated and compared in terms of learning rate to the one-step $Q$-Learning and Dyna-$Q$ algorithms for the same problem.

The experiment consists of searching for optimal paths, i.e., the shortest path with the lowest cost between two states. To study this problem in the context of reinforcement learning, we assume that it is a Markov decision process, where there is a set of possible states and a set of actions. A typical problem in path-finding is obstacle avoidance. The simplest approach to this problem is to ignore obstacles until jumping into them. This approach is simpler because it makes few demands: all that it needs is the relative position of the entity and its goal, and whether the immediate vicinity is blocked. For many game situations, this is good enough. But there are scenarios where the only intelligent approach would be to plan the entire route in advance.

In this paper, the playing space is represented with square tiles as a $39 \times 36$ grid (figure~\ref{fig:scenario}). The obstacles are walls that are set randomly (in gray). The state is the tile or position where the entity is located. Neighboring states would vary depending on the game and the local situation. The cost of going from one position to another can represent many things: in this case it is computed as the simple distance between the two positions, which in RL terminology is equivalent to set $r=-1$ for all non-terminal state transitions, minimizing thus the total distance, i.e. finding an optimal path. The grid is represented as a two dimensional matrix of 39 rows and 36 columns. This matrix establishes the communication between nodes or states; each node can be related up to four neighbors, depending on the type of each node, i.e. up $(\uparrow)$, down $(\downarrow)$, left $(\leftarrow)$ and right $(\rightarrow)$.

\section{Experimental Results}

\begin{figure}
\center
  \includegraphics[width=8.5 cm]{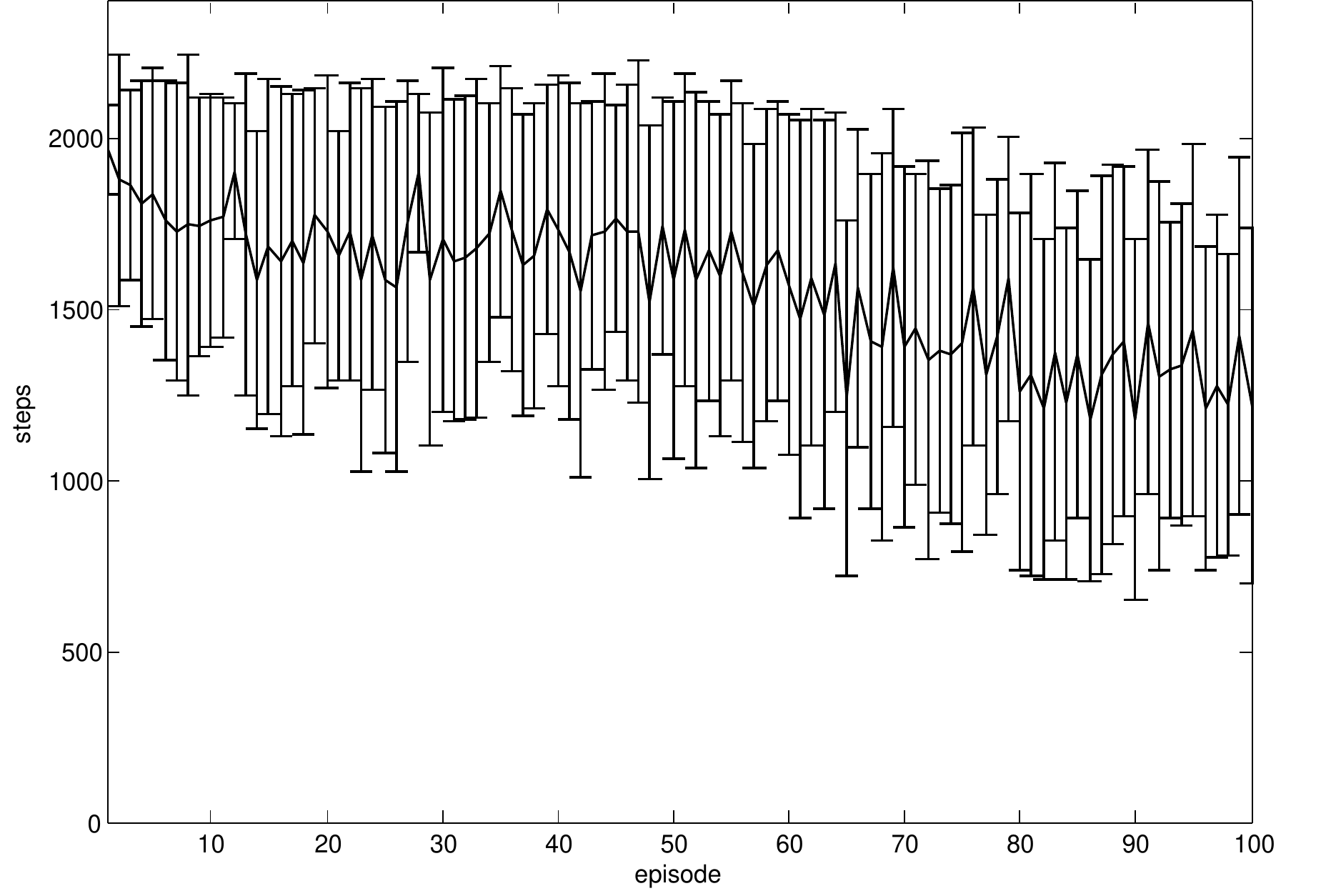}
  \caption{Average learning curve over 30 runs for the one-step tabular $Q$-Learning algorithm}\label{ql}
\end{figure}

\begin{figure}
\center
  \includegraphics[width=8.5 cm]{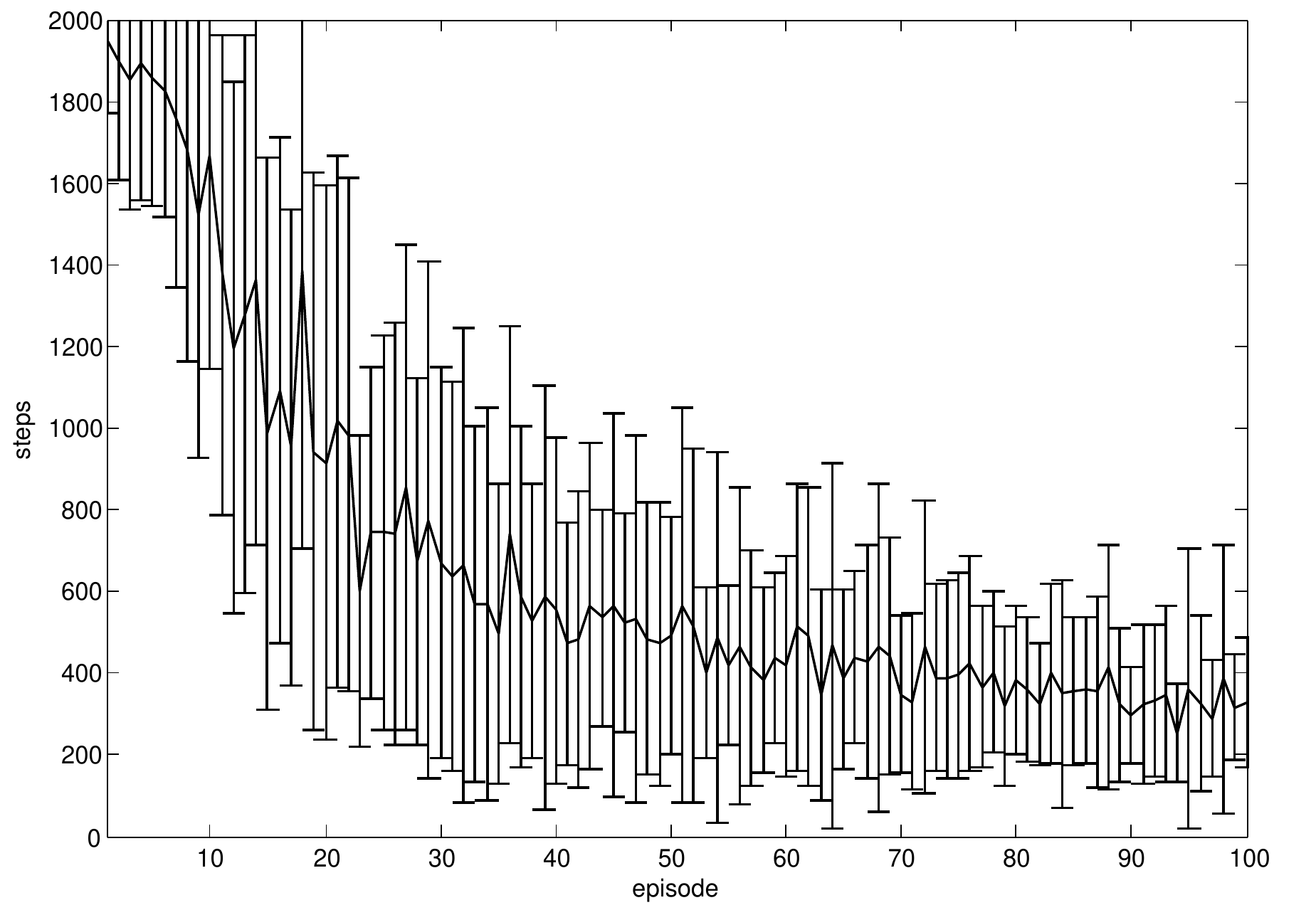}
  \caption{Average learning curve over 30 runs for the Dyna-$Q$ model with random sample with 10 planning steps}\label{dq}
\end{figure}

\begin{figure}
\center
  \includegraphics[width=8.5 cm]{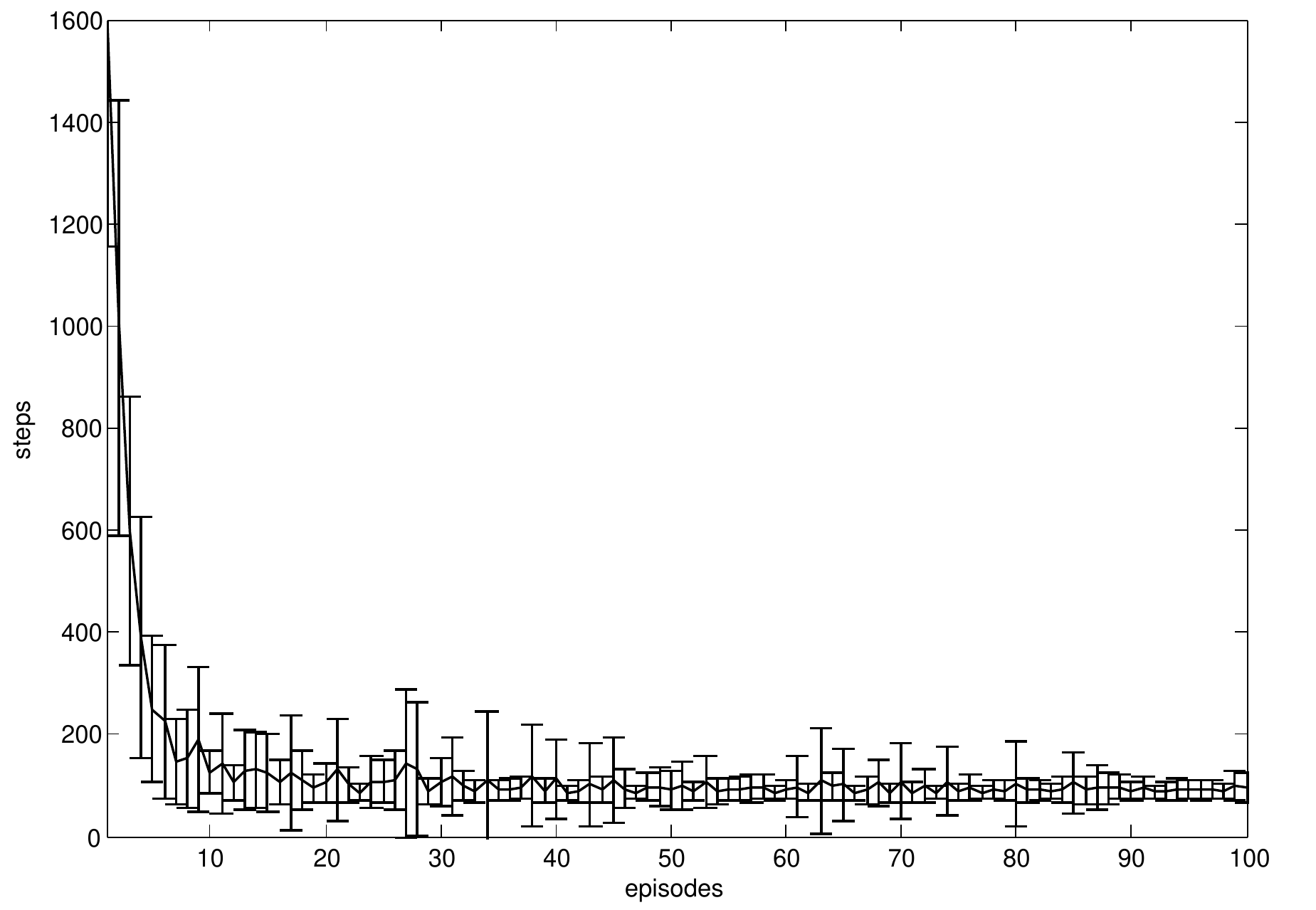}
  \caption{Average learning curve over 30 runs for the proposed Dyna-$\mathcal{H}$ heuristic planning algorithm with 10 planning steps}\label{hp}
\end{figure}

\begin{figure}
  \center
  \includegraphics[width=6.5 cm]{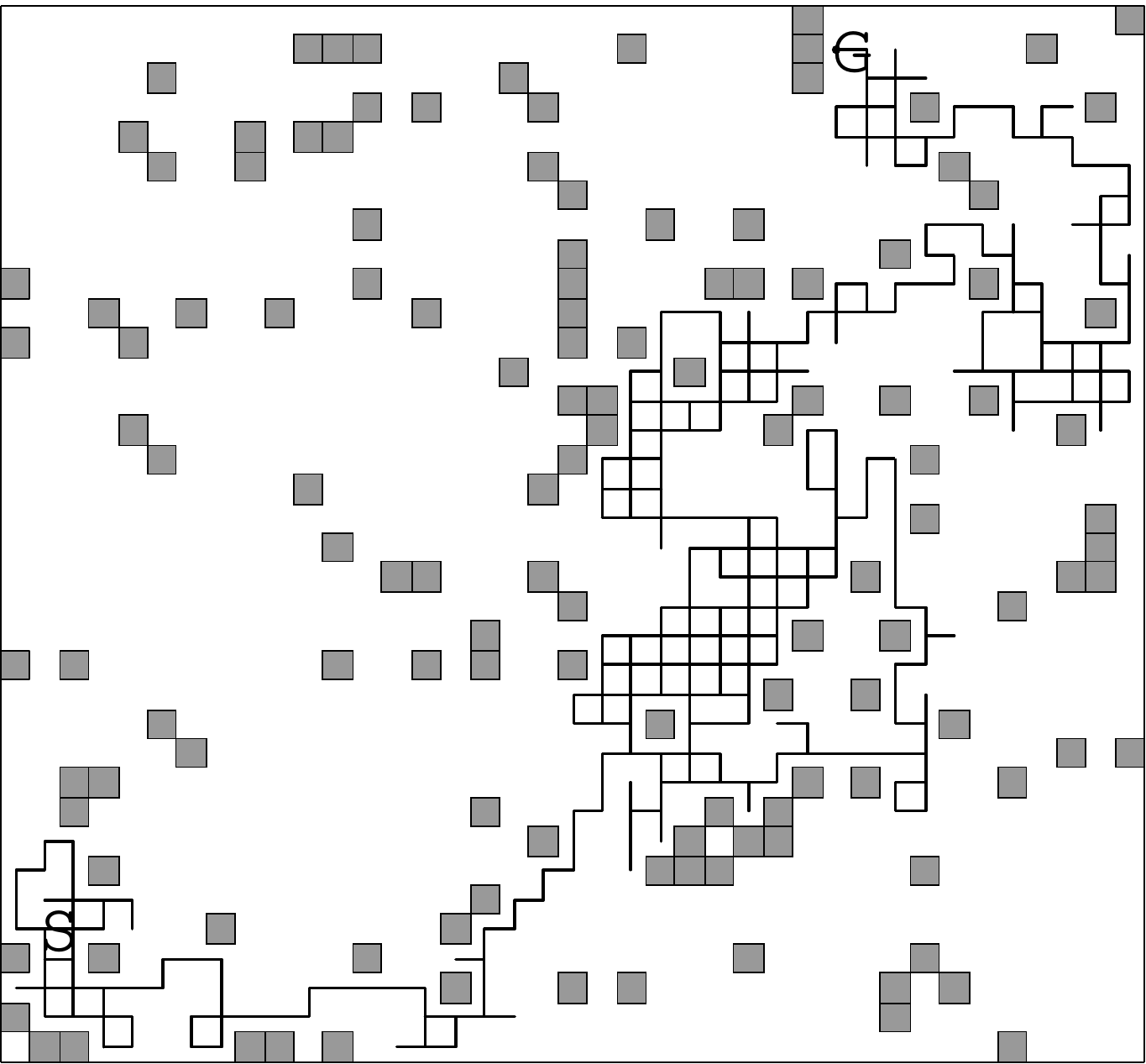}
  \caption{Trajectory describing the best path found by the one-step $Q$-Learning algorithm after 100 episodes for the first experiment.}\label{ql:path}
\end{figure}

\begin{figure}
  \center
  \includegraphics[width=6.5 cm]{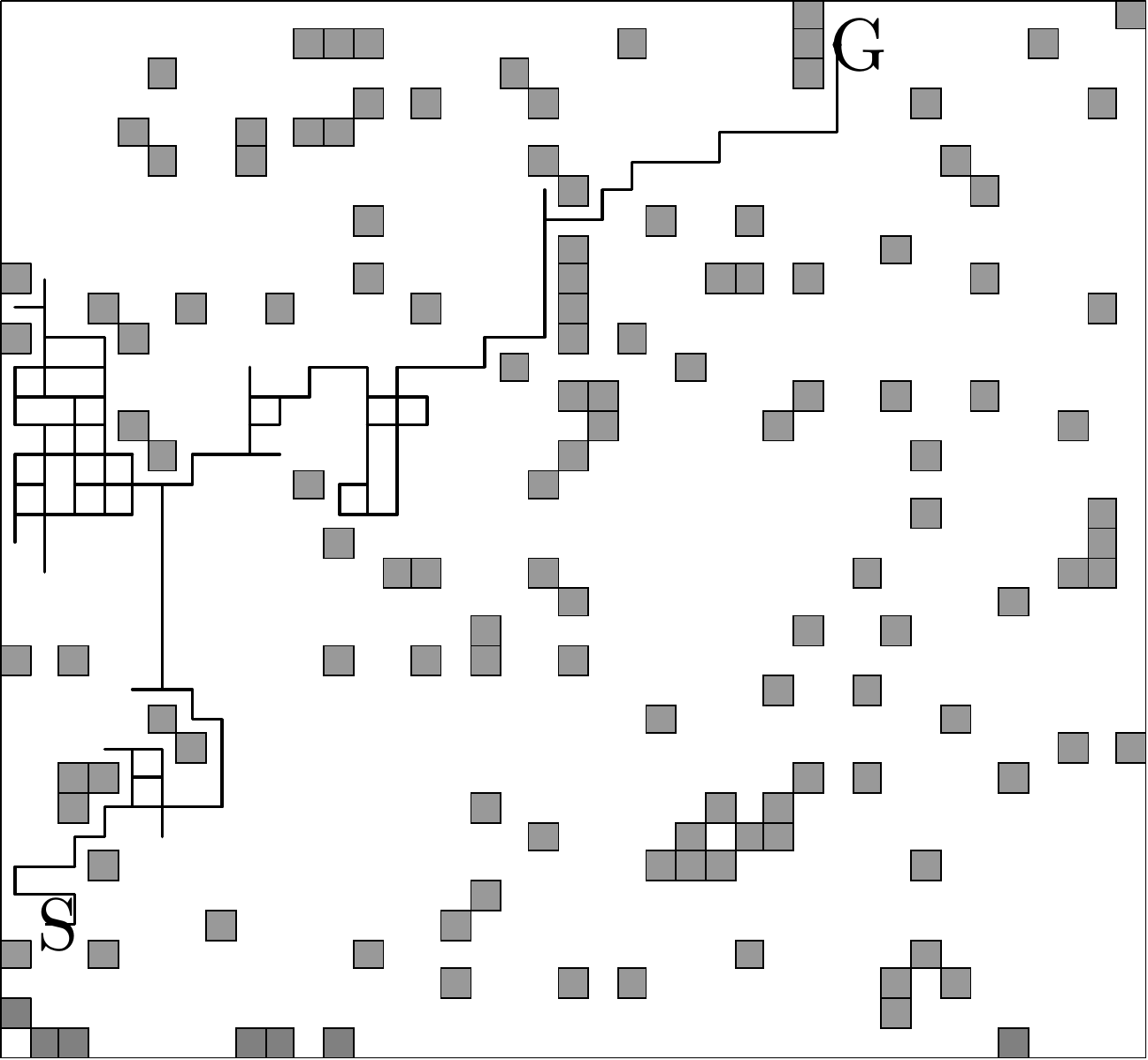}
  \caption{Trajectory describing the best path found by the Dyna-$Q$ planning algorithm (10 planning steps) after 100 episodes for the first experiment.}\label{dq:path}
\end{figure}

\begin{figure}
  \center
  \includegraphics[width=6.5 cm]{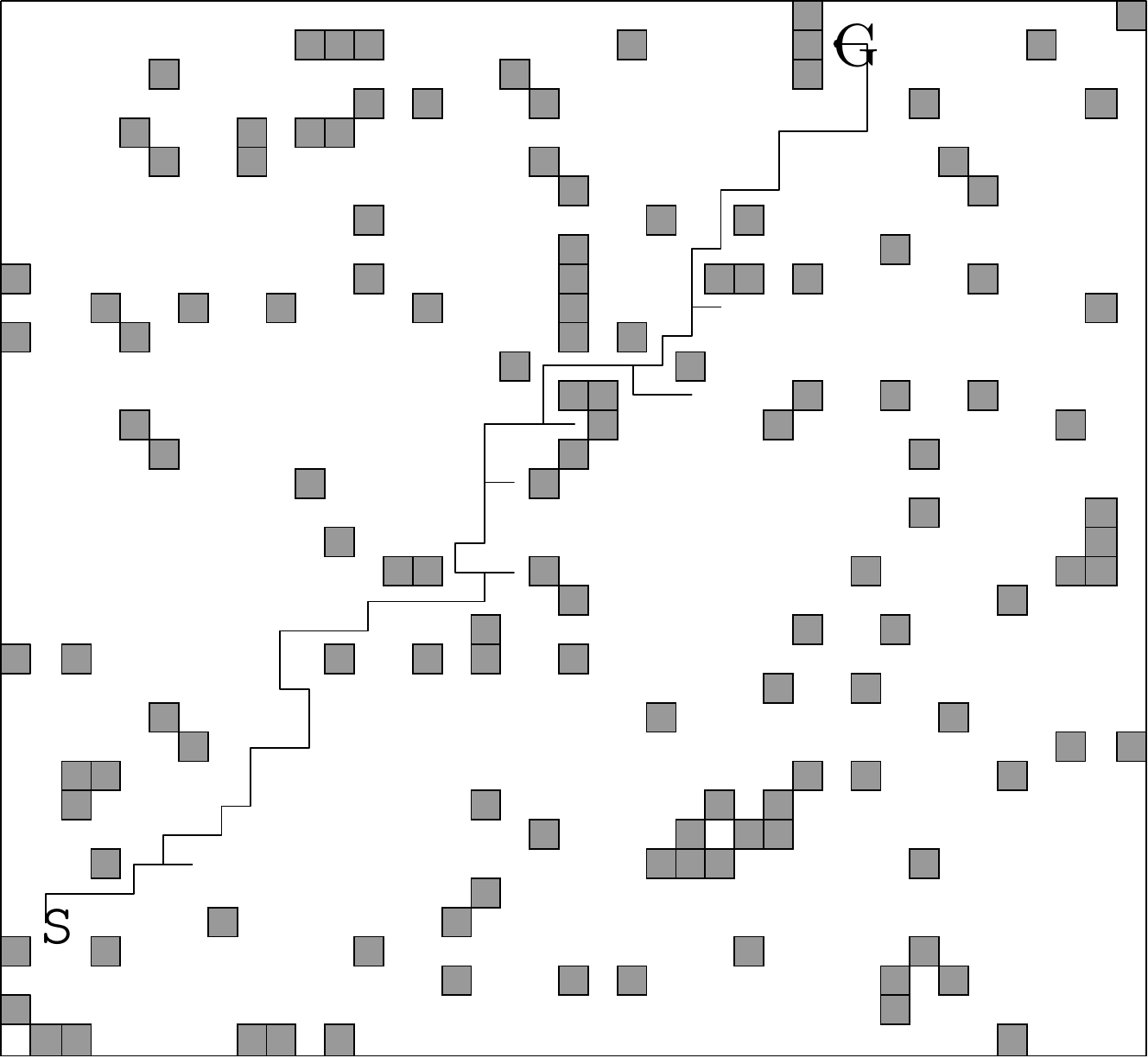}
  \caption{Trajectory describing the best path found by the proposed Dyna-$\mathcal{H}$ heuristic planning algorithm (10 planning steps) after 100 episodes for the first experiment.}\label{hp:path}
\end{figure}

\begin{figure*}
  \center
  \includegraphics[width=16 cm]{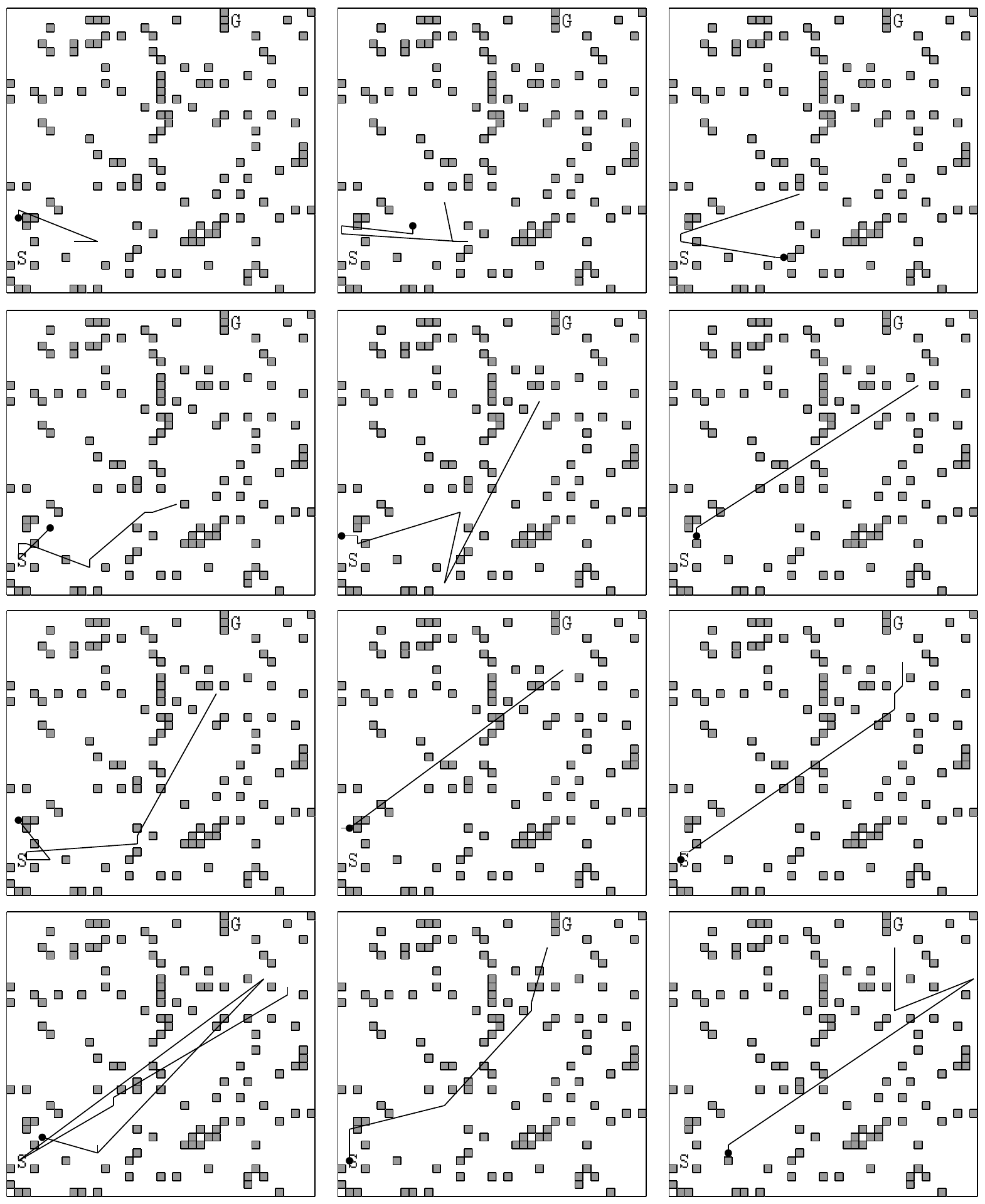}
  \caption{Several sampling trajectories produced by the heuristic sampling for consecutive time steps during one episode}\label{trajectories}
\end{figure*}

\begin{figure*}
  \center
  \includegraphics[width=10 cm]{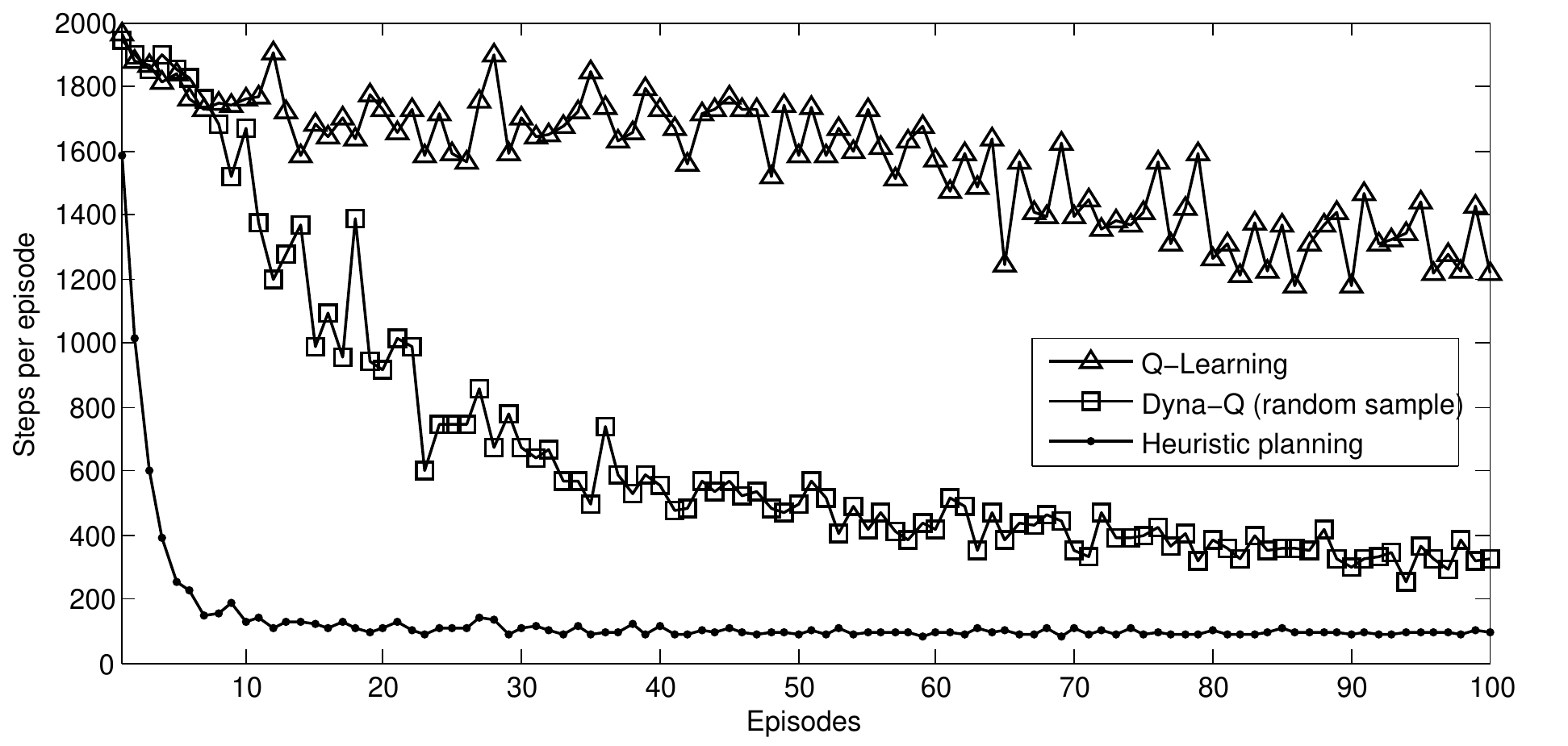}
  \caption{Comparison of the average learning curve over 30 runs for $Q$-Learning, Dyna-$Q$ (random sample with 10 planning steps) and the proposed Dyna-$\mathcal{H}$ heuristic planning algorithm (with 10 planning steps)}\label{comp}
\end{figure*}

Figures~\ref{ql} to \ref{comp} show the results of the simulations. As explained before, we have compared the performance of three algorithms: one-step $Q$-Learning (figures~\ref{ql} and ~\ref{ql:path}), Dyna-$Q$ (figures~\ref{dq} and ~\ref{dq:path}) and the proposed heuristic planning Dyna-$\mathcal{H}$ algorithm (figures~\ref{hp} and ~\ref{hp:path}).



As in Dyna maze~\citep{RL98}, all the tests were based on the one-step $Q$-Learning algorithm with a set of fixed parameters. The initial action values are zero, i.e. $Q(s,a)=0$, the step-size parameter is $\alpha=0.1$, and the exploration parameter was fixed to $\epsilon = 0.1$. When selecting greedily among actions, ties were broken randomly. For each algorithm, the learning curve shows the number of steps taken by the agent in each episode, averaged over 30 runs, each run consisting on a randomly generated labyrinth except from the staring ($1, 4$) and goal ($28, 34$) positions that remained constant during all experiments. Each random labyrinth was obtained using the same probability distribution (normal with $\mu=0, \sigma=0.3$) for every square tile of the grid, as shown in~(\ref{eq:random:tiles}).

\begin{align}
\phi(x) &=  \mathcal{N}(\mu=0,\,\sigma^2 = 0.3^2), \\
\label{eq:random:tiles} {tiletype} &= sgn\left( {abs}\left(   {Round}\left( \phi(x) \right)  \right)\right);
\end{align}
where ${tiletype}=1$ means that there is an obstacle and ${tiletype}=1$ indicates a free tile.

For each different algorithm, the initial seed for the random number generator was held constant, hence, all are evaluated on the same set of 30 different grid configurations. For Dyna-$Q$ and Dyna-$\mathcal{H}$, the number of planning steps was fixed to 10. All experiments ran for up to 100 episodes.

\begin{figure*}
  \center
  \includegraphics[width=10 cm]{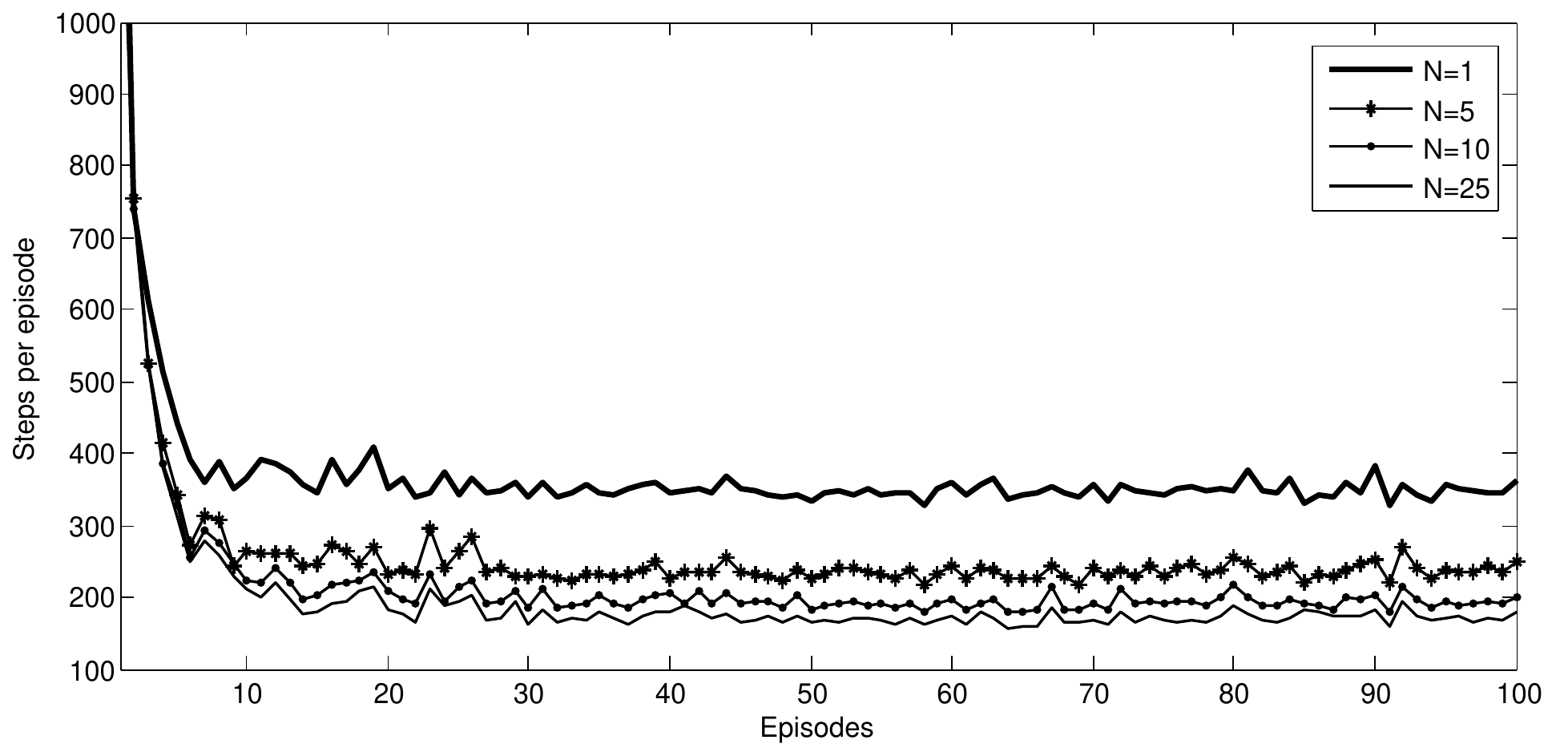}
  \caption{Average learning rates (over 30 runs) of the proposed Dyna-$\mathcal{H}$ heuristic planning algorithm for different numbers of planning steps, $N=$ 1, 5, 10 and 25}\label{ConvergevsP}
\end{figure*}

Figure~\ref{ql} shows the learning curve of the one-step $Q$-Learning algorithm. As it can bee seen, this is the slowest method and thus it serves as a standard for comparisons. The $Q$-Learning agent presents a very slow convergence curve and in fact it never found the optimal policy. It started with 2000 steps and showed a constant policy improvement during the 100 episodes, ending with approximately 1400 time steps. In figure~\ref{ql:path} the best path found by the one-step $Q$-Learning algorithm is shown.

Figure~\ref{dq} shows the learning rate of the Dyna-$Q$ algorithm. As expected, the Dyna agent improved the learning curve regarding the on-step $Q$-Learning algorithm. The Dyna-$Q$ agent presents a ``reasonable'' convergence curve. However, it never found the optimal policy. It started with 2000 steps and showed a high policy improvement up to episode 40, were the agent continued improving but with a slower rate, almost constant (linear like) factor during the remaining 60 episodes, ending the learning with around 400 time steps. Although it presents a good behavior, it could not found the optimal trajectory during the simulation time. Next we present some examples of the kind of solution trajectories generated by each algorithm. These solutions corresponds to the first experiment of each algorithm evaluation. In figure~\ref{dq:path} the best path found by Dyna-$Q$ algorithm is shown.

Figure~\ref{hp} shows the behavior of the proposed heuristic planning algorithm. As it is possible to see, the heuristic-planning agent improved a lot regarding the learning curve in comparison to the other algorithms. It presents an exponential convergence until the optimal policy is found. It started with 1600 steps and reduced them drastically up to episode 10, where it reaches the optimum (80 steps per episode). This means a high improvement both in the learning speed and the quality of the policy found. In figure~\ref{hp:path} the best path found by Dyna-$\mathcal{H}$ algorithm is shown. It can be seen that the generated path is very close to the optimal path.

Figure \ref{trajectories} shows several examples of the trajectories generated by the heuristic sampling planning procedure. The trajectories shown are all taken from the first episode of the first experiment of the Dyna-$\mathcal{H}$ algorithm and represent successive time steps of the episode. In these images, it is possible to appreciate clearly that the trajectories generated by the heuristic sampling strategy are almost the worst or very bad with respect to the solution, i.e. sampling from the worst trajectories, as defined by the Dyna-$\mathcal{H}$ algorithm and that using these trajectories the algorithm learns extremely well.

In figure~\ref{comp}, the average learning curves of the three algorithms are shown. The difference in terms of learning rate exhibited by the Dyna-$\mathcal{H}$ algorithm is evident.

As \citet{RL98} comment, in the short term, sampling according to, for instance, the on-policy distribution helps to focus on states that are close descendants of the initial state. On the other hand, in the long run, focusing on the on-policy distribution may make the convergence worse because the most visited states have already their correct values. Sampling them is useless, whereas sampling other states may actually help. This can be the reason why the exhaustive, unfocused approach, works better in the long run, at least for small problems. Although it may seem the same case, the proposed planning process does exactly the contrary to what would be an optimal policy (the policy to which the on-policy distribution should converge), focusing on apparently not very promising branches. However, by sampling from the worst trajectories, the learned policy converge quickly to the optimal one.

In figure~\ref{ConvergevsP} an analysis of the convergence of the proposed Dyna-$\mathcal{H}$ algorithm, for different numbers of planning steps $N$ is shown. The proposed heuristic planning algorithm have been tested for $N= 1, 5, 10$ and $25$ planning steps. For $N=1$, the algorithm converges in a few steps, around the 7th episode. However, it converges to a local suboptimal solution around 370 steps per episode. For $N=5$, the algorithm also converges in around 7 episodes but it converges to a suboptimal solution that is significantly better than for the previous case, reaching an average of 250 steps per episodes. The cases of $N=10$ and $N=25$ show an identical convergence pattern as the $N=5$ case but they reach better optimal policies.

It is quite significant that the case $N=1$ presents the same convergence rate than much higher planning rates, but it finds much worse policies. However, dealing with problems where the system should save computational resources, it can achieve a good compromise between optimality and computational time. The learning curves for $N=5$ up to $N=25$ are identical, being the only difference the optimality of the policy reached, that is, the length of the path from the initial node to the goal. Again, this behavior is quite interesting since it indicates that the trade off between optimality and computational resources can be directly controller by tuning the number of planning steps.

\section{Conclusions and further work}

In this paper we have presented a novel reinforcement learning-planning algorithm, Dyna-$\mathcal{H}$, that integrates planning and learning into an online algorithm based on the well known Dyna architecture. The proposed method involves heuristic in the planning module. It incorporates the ability of $A^*$ to focus on specific search subtrees in order to make the search more efficient by taking advantage of the heuristic. Besides, it is a model free strategy that can be applied to sequential decision making problems under uncertainty.

A scenario to compare three learning algorithms: $Q$-Learning, Dyna-$Q$ and the proposed Dyna-$\mathcal{H}$, has been designed. The results (learning rate and convergence and policy found) obtained by all these methods have been shown and discussed. The new algorithm gives the best trajectories and the number of steps is reduced in more than the 90\% with the Dyna-$\mathcal{H}$ strategy. From this results, we can conclude that the proposed Dyna-$\mathcal{H}$ heuristic planning algorithm is an effective strategy in path-finding problems and therefore for Role-Playing Games.

Since the main difference between Dyna-$Q$ and the proposed Dyna-$\mathcal{H}$ method is the use of a heuristic that guides the planning process when exploring the model, it makes sense to conclude that, under some well defined scenarios such as informed search methods, random sampling can be improved significatively.

We expect the successful application of the proposed algorithm to many related problems.

Further work should include the application of the proposed heuristic planning algorithm to different domains, for example, stochastic environments such as capture games for chaotic moving targets.

\section*{Software}
An open-source Matlab$^{TM}$ implementation of the Dyna-$\mathcal{H}$ algorithm can be obtained from the following direction: \url{http://www.dacya.ucm.es/jam/downloads/Dyna-H.rar}

\vspace{5mm}
\noindent \textbf{Acknowledgments.} This work is partially supported by Spanish Project DPI2009-14552-C02-01.

\bibliographystyle{elsarticle-harv}
\bibliography{references}

\end{document}